\documentclass[11pt]{article}

\usepackage[preprint]{acl}

\usepackage{times}
\usepackage{latexsym}
\usepackage{amsmath}
\usepackage{amssymb}
\usepackage{booktabs}
\usepackage{tablefootnote}
\usepackage{array}
\usepackage{pgfplots}
\usepackage{tikz}
\usepackage{enumitem}
\usetikzlibrary{positioning, arrows.meta}

\usepackage[T1]{fontenc}

\usepackage[utf8]{inputenc}

\usepackage{microtype}

\usepackage{inconsolata}

\usepackage{graphicx}

\title{Task-Conditional Flow Matching for Balanced Multilingual Text Embedding Adaptation}

\author{
 \textbf{Tirth Bhatt},
 \textbf{Naren Kumar S},
 \textbf{Mayank Singh}
\\
 LINGO Research Group, Indian Institute of Technology Gandhinagar, India
\\
 \small{
   \textbf{Correspondence:} \href{mailto:lingo@iitgn.ac.in}{lingo@iitgn.ac.in}
 }
}

\begin{document}
\maketitle
\begin{abstract}

Multilingual text embedding models are commonly adapted using a single training objective across diverse tasks, despite different tasks requiring fundamentally different optimization strategies. We introduce Task-Conditional Flow Matching (TCFM), a multilingual embedding adaptation framework that selectively applies Flow Matching to translation tasks while optimizing retrieval, classification, and pair-classification tasks with objectives better aligned to their learning dynamics. TCFM further combines teacher-guided representation preservation with a three-stage curriculum to enable stable adaptation. Evaluated on the Indic Massive Text Embedding Benchmark, TCFM establishes a new state-of-the-art, consistently improving embedding quality across a diverse set of multilingual tasks and generalizing across embedding model families. We will publicly release the codebase and datasets upon acceptance of the paper.

\end{abstract}

\section{Introduction}

Multilingual text embedding models provide a unified representation for a wide range of downstream tasks, including retrieval, bitext mining, semantic textual similarity (STS), clustering, classification, and natural language inference. While recent multilingual encoders demonstrate strong zero-shot capabilities~\citep{conneau-etal-2020-unsupervised, wang2024multilingual}, adapting them to new languages and domains remains challenging due to the substantial variation in downstream task characteristics~\citep{muennighoff-etal-2023-mteb}.  A natural response to this challenge is to adapt a strong multilingual encoder using a mixture of parallel corpora and supervised downstream datasets. However, optimizing a single  objective across heterogeneous task families often produces conflicting training 
signals~\citep{aghajanyan-etal-2021-muppet}, where improvements on some tasks come at the expense of others~\citep{wang-etal-2020-balancing}.

Contrastive learning is currently the standard approach for embedding adaptation, as it efficiently clusters related texts by pushing apart in-batch negatives \citep{gao2021simcse}. However, it struggles in diverse multi-task or multilingual settings as it treats all unpaired instances in a batch as strict negatives, they often suffer from the false-negative problem, where semantically similar sentences are incorrectly pushed apart \citep{chuang2020debiased}. Furthermore, this rigid separation can fragment the continuous embedding space, disrupting the smooth geometry needed to accurately model cross-lingual transformations \citep{ethayarajh2019contextual}.

Flow Matching~\citep{lipman2023flowmatching} learns smooth transformations between continuous representations by modeling a vector field rather than directly aligning embeddings. This makes it well suited for multilingual embedding adaptation, where translation pairs provide naturally corresponding representations. However, retrieval, classification, and pair-classification capture fundamentally different semantic relationships, making a uniform transport objective sub-optimal across these heterogeneous tasks.

In this work, we propose \textbf{Task-Conditional Flow Matching (TCFM)}, a multilingual embedding adaptation framework that applies Flow Matching only to translation-style sentence pairs while optimizing other task families with objectives better suited to their semantic characteristics. TCFM combines task-aware objective routing, teacher-guided representation preservation, and a three-stage curriculum to improve multilingual embeddings. We further show that applying Flow Matching uniformly across all task families provides no additional benefit, validating the proposed task-conditional design.

Our contributions are summarized as follows:
\begin{itemize}[nosep,noitemsep]

\item We introduce \textbf{Task-Conditional Flow Matching (TCFM)}, which selectively applies Flow Matching to translation-style supervision while routing other tasks to more appropriate objectives.

\item We demonstrate the effectiveness of TCFM for Indic multilingual text embedding adaptation across multiple model scales, achieving improvements of \textbf{5.45\%} and \textbf{2.72\%} on Indic MTEB over the Harrier-0.6B and Qwen3-Embedding-8B base models, respectively.

\end{itemize}

\section{Related Work}

\paragraph{Sentence and multilingual embeddings.}
Sentence-BERT~\citep{reimers-gurevych-2019-sentence} demonstrated the effectiveness of siamese networks
for learning sentence embeddings, while SimCSE~\citep{gao2021simcse} showed that contrastive learning can produce high quality sentence representations using minimal augmentation. For multilingual settings, models such as LASER~\citep{artetxe2019massively} and LaBSE~\citep{feng2022language} learn shared embedding spaces using large scale parallel corpora and translation-based objectives. More recent multilingual embedding models continue to rely primarily on contrastive objectives for cross-lingual alignment~\citep{wang2024multilingual, zhang-etal-2024-mgte, chen2024bge}. 


\paragraph{Contrastive learning and false negatives.}
Contrastive objectives based on InfoNCE~\citep{oord2018representation} have become the standard training paradigm for sentence embeddings. While highly effective, they assume that all non positive examples should be separated, an assumption that is often violated in multilingual and multi-task datasets where examples may share labels, intents, or semantic neighborhoods ~\citep{chuang2020debiased}. Prior work has therefore explored multi-positive objectives~\citep{khosla2020supervised} and false-negative mitigation~\citep{chuang2020debiased} strategies to improve representation learning. 


\paragraph{Flow Matching.}
It is a generative learning framework that models the transformation between two probability distributions by learning a continuous velocity field ~\citep{lipman2023flowmatching}. By regressing velocity along a predefined path, Flow Matching learns a vector field that transports representations between distributions. Unlike diffusion models, which learn to reverse a stochastic noising process, Flow Matching learns the transport dynamics directly, resulting in a deterministic and computationally efficient training objective~\citep{lipman2023flowmatching}. Although Flow Matching has primarily been studied for image and generative modeling tasks, its ability to learn smooth transformations makes it an attractive objective for multilingual embedding alignment, where parallel translations naturally define semantically aligned pairs. 

\

\noindent Unlike recent multilingual embedding approaches, which commonly optimize heterogeneous supervision using a unified contrastive objective~\citep{wang2024multilingual,su2023oneembedder}, the proposed framework applies different optimization objectives according to the semantic characteristics of each task family.



\section{The TCFM Framework}

Let $f_\theta(x)\in\mathbb{R}^d$ denote the sentence embedding produced by the encoder for an input text $x$, where the embedding is obtained by applying the pooling strategy native to the backbone architecture (e.g., mean pooling or end-of-sequence pooling) to the token-level representations. Additionally, let $f_0(\cdot)$ denote the corresponding frozen pretrained teacher encoder, which provides stable reference representations throughout adaptation. Each training instance consists of an anchor $x_i$, which serves as the primary reference text for the given task (e.g., a search query, a premise, or a source-language sentence). Associated with this anchor are a positive example $y_i$, an optional set of additional positives $P_i$, an optional hard negative $n_i$, and a task family label $r_i$. 


\subsection{Task Families}
\label{section:taskfamilies}
We curate a training data consisting of four task families as discussed below.

\begin{itemize}[nosep,noitemsep]

\item \textbf{Translation.}
consists of parallel sentences and semantically equivalent pairs 
which provide explicit source-target correspondence, making it well suited for learning continuous semantic transformations across languages.

\item \textbf{Classification.}
consists of samples that share the same label. Although these samples belong to the same semantic class, they are not necessarily paraphrases. 

\item \textbf{Pair Classification.}
Pair classification datasets provide labeled sentence pairs that describe the semantic relationship between two texts, such as whether they are semantically similar or belong to different relationship categories.

\item \textbf{Retrieval and Re-ranking.}
Retrieval and re-ranking datasets consist of query-document pairs, where the objective is to capture the relevance of candidate documents to an information need. Unlike translation pairs, query and document play distinct semantic roles and are therefore not interchangeable, making the relationship inherently asymmetric.

\end{itemize}


While the broader literature defines up to eight task families~\citep{muennighoff-etal-2023-mteb}, our framework categorizes tasks by their optimization compatibility. By mapping tasks with shared contrastive characteristics,such as clustering and semantic textual similarity into unified families, we ensure each group receives a coherent training objective. This structure readily adapts to new tasks matching their learning signals.

\subsection {Training Objectives}
\subsubsection{Multi-Positive Contrastive Loss}

Contrastive learning serves as a alignment objective for representation learning. Given a batch of texts $B$, the multi-positive InfoNCE objective for an anchor $x_i$ is formulated as:

\begin{equation}
\mathcal{L}_{\mathrm{cl}}(i) = -\log \frac{\sum_{p \in \{y_i\} \cup P_i} \exp\left(f_\theta(x_i)^\top f_\theta(p)/\tau\right)}{\sum_{c \in B} \exp\left(f_\theta(x_i)^\top f_\theta(c)/\tau\right)}
\end{equation}

where $\{y_i\} \cup P_i$ denotes the set of all positive candidates associated with anchor $x_i$, and $\tau$ is the temperature parameter.


We apply this objective symmetrically for translation (both languages as anchors) and asymmetrically for retrieval (query to document only). To mitigate false negative supervision from unlabeled in-batch matches during translation training, we exclude highly similar non-positives from the denominator using a similarity threshold. Conversely, this filtering is disabled for pair-classification, where similar negatives provide essential learning signals.

\subsubsection{Flow Matching}


Flow Matching models continuous transformations between probability distributions by learning a deterministic vector field. Given a source embedding $z_s = f_\theta(x_i)$ and a target embedding $z_{tgt} = f_\theta(y_i)$, we sample an interpolation time $t\sim U(0,1)$ and construct the intermediate representation
 \begin{equation}
z(t)=(1-t)z_s+t z_{tgt}.
\end{equation}

The corresponding target velocity is
\begin{equation}
u=z_{tgt}-z_s.
\end{equation}

To stabilize optimization, the target velocity can be interpolated with the corresponding translation direction predicted by the frozen teacher encoder:
\begin{equation}
u^\star
=
(1-\alpha)(z_{tgt}-z_s)
+
\alpha\left(f_0(y)-f_0(x)\right),
\end{equation}
where $\alpha$ controls the contribution of the teacher-guided transport direction.

The velocity prediction network $v_\phi(\cdot)$ is conditioned on the task instruction embedding extracted from the frozen teacher encoder. Let $c=f_0(I_r)$, where $I_r$ denotes the task-specific instruction associated with task family $r$. The Flow Matching objective minimizes the discrepancy between the predicted and target transport directions:
\begin{equation}
\mathcal{L}_{flow}
=
1-
\cos\!\left(
v_\phi(z(t),t,c),
u^\star
\right).
\end{equation}

The Flow Matching objective learns a local transport field but does not necessarily update the encoder representations. To allow the learned vector field to influence the embedding space while preventing large representation shifts, we introduce a bounded transport objective. The transported embedding is computed as
\begin{equation}
\hat{z}
=
\operatorname{norm}
\left(
z_s
+
\eta\,
\operatorname{norm}
\left(
v_\phi(z_s,0,c)
\right)
\right),
\end{equation}
where $\eta$ is a small transport step.

The transported representation is then encouraged to align with the target embedding:
\begin{equation}
\mathcal{L}_{transport}
=
1-
\cos
\left(
\hat{z},
\operatorname{stopgrad}(z_{tgt})
\right).
\end{equation}

Parallel translation pairs represent semantically equivalent sentences across languages and therefore admit meaningful continuous transformations in the embedding space. However, retrieval, classification, and pair-classification tasks optimize fundamentally different semantic relationships and do not naturally define transport trajectories between examples. Consequently, in our framework, Flow Matching is activated only for translation task families. Together, the flow matching and bounded transport objectives regularize multilingual alignment during translation training, while allowing the remaining task families to retain optimization objectives that better reflect their supervision characteristics.

\begin{table*}[t]
\centering
\small
\renewcommand{\arraystretch}{1.15}
\begin{tabular}{>{\raggedright\arraybackslash}p{3.0cm}
>{\raggedright\arraybackslash}p{2.2cm}
>{\raggedright\arraybackslash}p{5.2cm}
c c c}
\toprule
\textbf{Dataset} &
\textbf{Task Family} &
\textbf{Description} &
\textbf{Total} &
\textbf{Train} &
\textbf{Languages} \\
\midrule

\texttt{local\_wide\_parallel}
&
Translation
&
English--Indic parallel sentence pairs collected for multilingual translation alignment.
&
7.9M &
130k &
16 Indic + En
\\

\texttt{Samanantar}~\citep{ramesh2022samanantar}
&
Translation
&
Large-scale English--Indic parallel corpus containing semantically equivalent sentence pairs.
&
49.7M &
55k &
11 Indic + En
\\

\texttt{MASSIVE Same Intent}~\citep{fitzgerald2022massive}
&
Classification
&
Multilingual utterances sharing the same intent label within each language.
&
1M &
60k &
8 Indic + En
\\

\texttt{MASSIVE Aligned Intent}~\citep{fitzgerald2022massive}
&
Classification
&
Cross-lingual intent dataset aligning semantically equivalent utterances across languages.
&
1M &
50k &
8 Indic + En
\\

\texttt{IndicSentiment} \footnotemark[1]

&
Classification
&
Sentence-level sentiment classification dataset annotated with positive, negative, and neutral labels.
&
~28k &
20k &
14 Indic
\\

\texttt{IndicXNLI}~\citep{aggarwal2022indicxnli}
&
Pair Classification
&
Cross-lingual natural language inference dataset with entailment, contradiction, and neutral sentence pairs.
&
4.4M &
5k &
11 Indic
\\

\texttt{IndicMSMARCO}~\citep{prasanjith2025indicragsuite}
&
Retrieval
&
Multilingual query--document relevance dataset for passage retrieval.
&
11.4M &
24k &
12 Indic
\\

\texttt{Bhasha}~\citep{madhani2023bhasa}
&
Language Identity~\footnotemark[2]
&
Monolingual corpus spanning Indic languages used for language identity preservation.
&
123k &
34k &
22 Indic
\\

\bottomrule
\end{tabular}
\caption{Training datasets used by TCFM. The ``Total'' column denotes the full available scale of the source dataset, while the ``Training'' column reports the number of instances sampled for our multi-stage curriculum.}
\label{tab:training_data}
\end{table*}

\subsubsection{Teacher Preservation}

Unconstrained optimization can gradually drift away from the semantic structure learned by the pretrained encoder. To preserve this prior knowledge, we regularize the adapted model using a frozen teacher encoder $f_0$ through both pointwise and relational objectives.

The pointwise teacher loss encourages each example to remain close to its original embedding direction:
\begin{equation}
\mathcal{L}_{teacher} = 1 - \cos(f_\theta(x_i), f_0(x_i))
\end{equation}
where $f_\theta(\cdot)$ and $f_0(\cdot)$ denote the representations produced by the student and frozen teacher, respectively. While the pointwise objective preserves individual embeddings, it does not constrain relationships between examples. We therefore introduce a relational preservation objective that matches the pairwise similarity structure of each mini-batch:
\begin{equation}
\mathcal{L}_{rel} = \left\| H_\theta H_\theta^\top - H_0 H_0^\top \right\|_F^2
\end{equation}
where $H_\theta$ and $H_0$ denote the matrices formed by stacking the $L_2$-normalized sentence representations of all examples in the mini-batch produced by the student and teacher encoders, respectively. 

Flow Matching and contrastive learning aggressively adapt the embedding space toward the target multilingual tasks. We apply these dual preservation objectives to regularize this adaptation process. Together, these constraints ensure that the adapted model retains the global semantic relationships learned by the pretrained multilingual encoder while effectively specializing in the new task families.

\footnotetext[1]{\url{https://huggingface.co/datasets/ai4bharat/IndicSentiment}}
\footnotetext[2]{This dataset is not a part of our task family, but rather used for monolingual consistency}

\subsubsection{Hard-Negative Margin Repair}

Standard contrastive objectives can inadvertently reduce the separation between semantically related but distinct examples~\citep{wang2021understanding}. To enforce a strict discriminative boundary, a cosine-margin objective~\citep{reimers-gurevych-2019-sentence} is employed to ensure that an anchor $x$ remains closer to its positive sample $y$ than to a hard negative $n$ by at least a predefined margin:

\begin{equation}
\begin{aligned}
\mathcal{L}_{hn}
=
\max\Big(
0,\,
m
+ 
\cos(f_\theta(x_i),f_\theta(n_i))
- \\
\cos(f_\theta(x_i),f_\theta(y_i))
\Big), 
\end{aligned}
\end{equation}
where $m$ is the margin hyperparameter. In our framework, certain task families, particularly natural language inference and retrieval, provide explicit hard-negative examples that require this strict separation. Consequently, this objective is activated only for task families that contain explicit hard negatives. By conditionally applying this loss, the margin repair mechanism complements the primary contrastive objective, preserving the relative ordering between positive and negative examples without affecting the continuous translation-style alignment.

\subsubsection{Monolingual Consistency}

While cross-lingual alignment encourages semantically equivalent sentences from different languages to occupy nearby regions of the embedding space, excessive alignment may reduce the quality of monolingual representations. To preserve within-language semantic consistency, we incorporate a SimCSE-style self-consistency objective by encoding the same input twice under stochastic training conditions:

\begin{equation}
\mathcal{L}_{mono} = \mathcal{L}_{sym}\left(f_\theta(x_i), f_\theta(\tilde{x}_i)\right)
\end{equation}

where $\tilde{x}$ denotes a second encoding of the same input obtained under independent dropout masks during training. This objective regularizes local neighborhoods and improves the robustness of monolingual representations without requiring additional supervision. The cross-lingual alignment encourages semantically equivalent sentences from different languages to occupy nearby regions of the embedding space, excessive alignment can inadvertently reduce the quality of monolingual representations. To preserve within-language semantic consistency, we incorporate this SimCSE-style self-consistency objective. It regularizes local neighborhoods and ensures the monolingual integrity of the representations remains robust as cross-lingual clusters are merged, all without requiring additional supervision.

\subsubsection{Overall Objective and Curriculum}

The complete training objective combines the task-specific losses introduced above:

\begin{equation}
\begin{aligned}
\mathcal{L}
={}&
\lambda_c\mathcal{L}_{cl}
+
\left(
\lambda_f\mathcal{L}_{flow}
+
\lambda_t\mathcal{L}_{transport}
\right)
\\
&
+
\left(
\lambda_p\mathcal{L}_{teacher}
+
\lambda_r\mathcal{L}_{rel}
\right)
\\
&
+
\lambda_h\mathcal{L}_{hn}
+
\lambda_m\mathcal{L}_{mono}.
\end{aligned}
\end{equation}

where each $\lambda$ controls the contribution of its corresponding objective.


\subsection{Three-Stage Training Curriculum}
\label{sec:threestagecurriculum}

TCFM is trained using a three-stage curriculum designed to progressively introduce increasingly diverse supervision while preserving the multilingual representations learned during earlier stages.

\paragraph{Stage 1: Cross-Lingual Alignment.}
The first stage focuses exclusively on translation-style supervision using the parallel corpora listed in Table~\ref{tab:training_data}. During this stage, Flow Matching and teacher-preservation objectives receive greater emphasis, enabling the model to establish stable multilingual representations before introducing more heterogeneous supervision.

\paragraph{Stage 2: Multi-Task Semantic Adaptation.}

The second stage introduces multilingual classification and natural language inference datasets while maintaining a replay buffer of approximately 40k translation pairs from Stage~1. By ensuring that this replay data constitutes approximately 23\% of the overall training mixture, we effectively mitigate the catastrophic forgetting of the cross-lingual alignment established during the first stage \citep{rolnick2019experience}. Throughout this phase, we also progressively increase the contribution of contrastive learning and hard-negative objectives.

\paragraph{Stage 3: Retrieval Adaptation and Monolingual Regularization.}

The final stage incorporates asymmetric query-document retrieval together with monolingual language-identity supervision. The 40,000-sample replay buffer from the previous stages is retained throughout this training phase, comprising approximately 41\% of the total Stage~3 data. Concurrently, the monolingual consistency objective regularizes within-language representations without sacrificing the cross-lingual alignment learned during the earlier stages.


\section{Experimental Setup}

\subsection{Training Data}
\label{sec:trainingdata}

We construct a balanced multi-task training mixture across the four task families (Table~\ref{tab:training_data}), converting each source dataset into its objective-specific supervision format, ranging from symmetric parallel pairs for translation and class-based semantic pairs for classification, to explicit entailment-contradiction pairs for NLI and asymmetric query-document pairs for retrieval. To enrich continuous cross-lingual alignment, our translation dataset incorporates \texttt{local\_wide\_parallel}, a self-curated corpus constructed by translating English Wikipedia sentences\footnote[3]{\url{https://huggingface.co/datasets/sentence-transformers/wikipedia-en-sentences}} into 16 Indic languages using Sarvam-Translate\footnote[4]{\url{https://huggingface.co/sarvamai/sarvam-translate}}. To prevent gradient imbalance~\citep{wang-etal-2020-balancing} caused by massive parallel corpora (e.g., Samanantar) overpowering low-resource tasks, we employ stratified subsampling across languages and cap maximum dataset sizes rather than preserving native distributions. Extended details on dataset formatting, preprocessing pipelines, and sampling strategies are provided in Appendix~\ref{sec:appendix_data_details}.

\begin{table*}[t]
\centering
\small
\renewcommand{\arraystretch}{1.15}
\setlength{\tabcolsep}{2pt}
\resizebox{\textwidth}{!}{%
\begin{tabular}{l ccc ccc ccc ccc ccc}
\toprule
&
\multicolumn{3}{c}{\textbf{Gemma-300M}} &
\multicolumn{3}{c}{\textbf{Harrier-270M}} &
\multicolumn{3}{c}{\textbf{Harrier-0.6B}} &
\multicolumn{3}{c}{\textbf{Qwen-4B}} &
\multicolumn{3}{c}{\textbf{Qwen3-8B}} \\
\cmidrule(lr){2-4}
\cmidrule(lr){5-7}
\cmidrule(lr){8-10}
\cmidrule(lr){11-13}
\cmidrule(lr){14-16}
\textbf{Task Category} &
\textbf{Base} & \textbf{TCFM} & \textbf{$\Delta$} &
\textbf{Base} & \textbf{TCFM} & \textbf{$\Delta$} &
\textbf{Base} & \textbf{TCFM} & \textbf{$\Delta$} &
\textbf{Base} & \textbf{TCFM} & \textbf{$\Delta$} &
\textbf{Base} & \textbf{TCFM} & \textbf{$\Delta$} \\
\midrule
Bitext Mining & 60.80 & 65.28 & +4.48 & 69.51 & 72.69 & +3.18 & 76.94 & 78.51 & +1.57 & 75.67 & 76.70 & +1.03 & 77.61 & 78.36 & +0.75 \\
Classification & 64.34 & 66.71 & +2.36 & 62.99 & 68.03 & +5.05 & 64.58 & 67.85 & +3.27 & 69.59 & 71.00 & +1.41 & 70.62 & 72.39 & +1.77 \\
Clustering & 31.90 & 37.62 & +5.72 & 30.57 & 46.04 & +15.47 & 29.09 & 50.12 & \textbf{+21.03} & 39.74 & 51.45 & +11.71 & 40.50 & 55.75 & \textbf{+15.25} \\
Pair Classification & 68.06 & 68.30 & +0.24 & 63.02 & 67.96 & +4.94 & 63.77 & 67.72 & +3.95 & 77.08 & 78.95 & +1.87 & 81.24 & 82.08 & +0.84 \\
Reranking & 88.04 & 88.14 & +0.10 & 85.68 & 85.72 & +0.04 & 85.54 & 85.67 & +0.13 & 86.36 & 86.48 & +0.12 & 87.08 & 87.59 & +0.51 \\
Retrieval & 80.65 & 81.12 & +0.47 & 80.97 & 80.72 & -0.26 & 80.69 & 81.72 & +1.03 & 88.80 & 88.80 & +0.00 & 93.50 & 93.22 & -0.28 \\
STS & 43.74 & 46.27 & +2.53 & 42.64 & 43.73 & +1.09 & 48.82 & 51.06 & +2.24 & 51.69 & 51.98 & +0.29 & 59.45 & 60.10 & +0.65 \\
\midrule
\textbf{Average} & \textbf{64.34} & \textbf{66.68} & \textbf{+2.34} & \textbf{63.94} & \textbf{68.33} & \textbf{+4.40} & \textbf{65.87} & \textbf{69.46} & \textbf{+3.59} & \textbf{70.94} & \textbf{72.59} & \textbf{+1.65} & \textbf{72.89} & \textbf{74.87} & \textbf{+1.98} \\
\bottomrule
\end{tabular}
}
\caption{Task-category Indic MTEB results of TCFM across the evaluated model architectures.}
\label{tab:detailed_results}
\end{table*}

\subsection{Baseline Models}

We evaluate the proposed framework across multiple architectures. Evaluation studies are conducted using \texttt{google/embeddinggemma-300m} \citep{embedding_gemma_2025}, \texttt{microsoft/harrier-oss-v1-270m}, \texttt{microsoft/harrier-oss-v1-0.6B} and \texttt{Qwen/Qwen3-Embedding-4B}, while our best-performing model is obtained by adapting \texttt{Qwen/Qwen3-Embedding-8B} \citep{qwen3embedding2025} using LoRA. 

For Gemma-based models, sentence representations are obtained through mean pooling, whereas Harrier OSS and Qwen models employ left-padding safe EOS pooling. All embeddings are L2-normalized. The Flow Matching velocity network consists of a two-layer MLP with SiLU activations, Layer Normalization~\citep{ba2016layer}, and sinusoidal timestep embeddings, further architectural hyperparameters are detailed in Appendix~\ref{sec:appendix_reproducibility}. Although memory queues~\citep{cao-etal-2022-exploring} were explored during preliminary experiments, they were omitted from the final training recipe because they increased the likelihood of false negatives in heterogeneous multi-task batches.


\subsection{Evaluation Benchmark}

We evaluate the proposed framework on the Indic Massive Text Embedding Benchmark (Indic MTEB) \citep{enevoldsen2025mmtebmassivemultilingualtext}. Indic MTEB evaluates embedding quality on Bitext Mining, Semantic Textual Similarity (STS), Classification, Clustering, Pair Classification, Retrieval, and Reranking. Although Multilingual MTEB covers a broader collection of languages and tasks, Indic MTEB retains a diverse set of embedding task families while encompassing 25 linguistically diverse languages, making it a comprehensive benchmark for evaluating multilingual embedding models and the generalization of the proposed training objectives.

\subsection{Evaluation Protocol}

Our training follows the three-stage curriculum described in Section~\ref{sec:threestagecurriculum}. To analyze the contribution of each stage, we evaluate checkpoints obtained at the end of every curriculum stage in addition to the final model. Unless otherwise specified, all reported improvements are measured relative to the corresponding frozen base model.

\section{Results}
\label{sec:results}

We evaluate the TCFM framework on the Indic MTEB v3, spanning seven task families: Bitext Mining, Classification, Clustering, Pair Classification, Retrieval, Reranking, and Semantic Textual Similarity (STS).

We evaluate TCFM on both \texttt{microsoft/harrier-oss-v1-0.6b} and \texttt{Qwen/Qwen3-Embedding-8B} to analyze the behavior of the proposed training framework in controlled settings and its effectiveness on a substantially larger embedding model.
\begin{table*}[t]
\centering
\small
\begin{tabular}{lcccccccc}
\toprule
\textbf{Configuration} & \textbf{Macro ($\Delta$)} & \textbf{Bitext} & \textbf{Classif.} & \textbf{Cluster} & \textbf{PairCls.} & \textbf{Rerank} & \textbf{Retrieval} & \textbf{STS} \\
\midrule
Base Model & 65.87 & 76.94 & 64.58 & 29.09 & 63.77 & 85.54 & 80.69 & 48.82 \\
\midrule
Contrastive & +1.96 & -2.38 & +2.91 & +20.30 & +4.45 & +0.36 &-6.08 & -3.83\\
Contrastive + Teacher & +2.81 & +0.17 & +2.26 & \textbf{+21.32} & +6.27 & +0.48 & +0.11 & +0.46 \\
Flow + Teacher & +1.67 & -0.20 & +0.73 & +21.15 & +2.58 & -0.05 & \textbf{+1.06} & -0.82 \\
Flow + Teacher (All Tasks) & +1.63 & -0.43 & +0.66 & +20.97 & +3.02 & +0.06 & +1.17 & -0.91 \\
Flow + Contrastive & +1.19 & -0.95 & +3.25 & +16.82 & +6.41 & -2.63 & -12.48 & -8.99 \\
Reordered Curriculum$^{\dagger}$ & +2.61 & +0.67 & +2.27 & +11.27 & \textbf{+7.41} & \textbf{+0.94} & +0.95 & +2.16 \\
Flow + Contrastive + Teacher & \textbf{+3.59} & \textbf{+1.57} & \textbf{+3.27} & +21.03 & +3.95 & +0.13 & +1.03 & \textbf{+2.24} \\
\bottomrule
\end{tabular}
\caption{Component-wise ablation study on the Harrier-0.6B encoder. Positive values denote improvements over the frozen base model.Results are reported as the average improvement for each MTEB task category, using the same category-level aggregation as per the Table \ref{tab:detailed_results}.
$^{\dagger}$ Stage Order Ablation denotes the curriculum Stage~1 $\rightarrow$ Stage~3 $\rightarrow$ Stage~2, instead of the proposed stages}.
\label{tab:harrier_06b_ablation}
\end{table*}

We observe an improvement in the Indic MTEB Score of ~\texttt{microsoft/harrier-oss-v1-0.6b}
by \textbf{+3.59} points.
The largest improvement is observed on the Clustering benchmark, where TCFM increases performance by more than 21 points while also improving performance on the remaining task families. 
Table~\ref{tab:detailed_results} summarizes category-level performance across models, while a fine-grained, per-task breakdown is provided in Appendix~\ref{sec:appendix_extended_evals}.
Cross-lingual alignment also benefits from the proposed framework, with \texttt{IN22ConvBitextMining} improving by +3.30 points and \texttt{IndicCrosslingualSTS} improving by +2.24 points.

Performance improvements are not uniform across every task category. While most categories benefit from adaptation, a small number of tasks exhibit a modest decrease, including \texttt{SanskritShlokasClassification} (-5.31), \texttt{IN22GenBitextMining} (-0.16), \texttt{NepaliNewsClassification} (-0.14), and \texttt{XQuADRetrieval} (-0.23). Despite substantial overall gains, multilingual embedding adaptation remains a challenging multi-objective optimization problem. 

To evaluate whether the proposed training strategy generalizes to larger language models, we apply the framework to the \texttt{Qwen3-Embedding-8B} using LoRA adaptation. Relative to the frozen base model,  an absolute gain of \textbf{+1.97} points is observed for the Indic MTEB. Improvements are observed across six of the seven benchmark categories (see Table ~\ref{tab:detailed_results}), while Retrieval exhibits a minor decrease (-0.28), indicating that the proposed framework maintains balanced performance while substantially improving performance in the remaining task families. Although the largest improvements are observed on the Clustering and Classification tasks. Cross-lingual alignment also benefits from adaptation, with conversational bitext mining improving by +1.32 points while maintaining competitive performance on generalized bitext mining and semantic textual similarity.

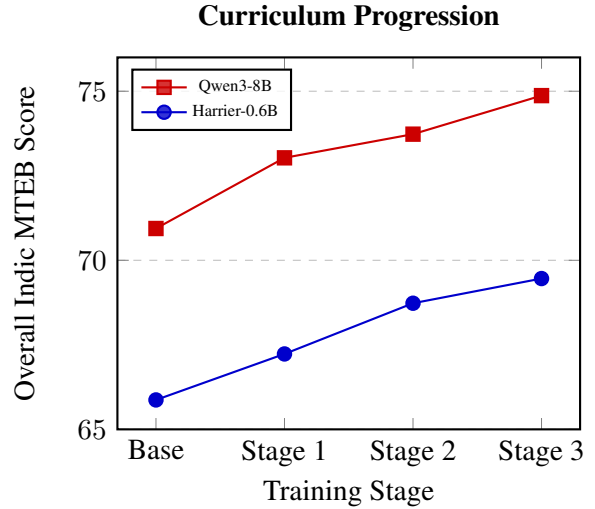
\begin{figure}[!tbh]
\centering
\begin{tikzpicture}
\begin{axis}[
    width=\columnwidth, 
    height=6.5cm,      
    title={\textbf{Curriculum Progression}},
    xlabel={Training Stage},
    ylabel={Overall Indic MTEB Score},
    xtick={0,1,2,3},
    xticklabels={Base, Stage 1, Stage 2, Stage 3},
    ymin=65, ymax=76,  
    legend pos=north west,
    legend style={font=\scriptsize, nodes={scale=0.9, transform shape}, fill=white, fill opacity=0.8, draw opacity=1, text opacity=1},
    ymajorgrids=true,
    grid style=dashed,
    thick,
    mark size=2.5pt
]

\addplot[color=red!80!black, mark=square*, mark options={fill=red!80!black}]
    coordinates {(0,70.94) (1,73.03) (2,73.73) (3,74.87)};
\addlegendentry{Qwen3-8B}

\addplot[color=blue!80!black, mark=*, mark options={fill=blue!80!black}]
    coordinates {(0,65.87) (1,67.23) (2,68.73) (3,69.46)};
\addlegendentry{Harrier-0.6B}

\end{axis}
\end{tikzpicture}
\caption{Step-wise performance trajectory across the TCFM training curriculum.}
\label{fig:curriculum_plot_dual}
\end{figure}

\section{Ablation Experiments}
\label{sec:ablations}

\subsection{Component Analysis}
To better understand the respective roles of individual components of TCFM. We perform ablation studies on the \texttt{Harrier-0.6B} which provides an appropriate balance between model capacity and the ability to systematically evaluate multiple training variants under a consistent experimental protocol$^{\dagger}$. Allowing us to study the individual effects of contrastive learning, teacher preservation, Flow Matching, and task-conditional routing. 
{\renewcommand{\thefootnote}{}\footnote{$^{\dagger}$ Unless otherwise specified, each experiment modifies a single component while keeping the remaining components.}}

Table~\ref{tab:harrier_06b_ablation} summarizes the contribution of the principal components of TCFM. Contrastive learning combined with teacher preservation yields strong improvements on Pair Classification (+6.27), Clustering (+21.32), and STS (+0.46), while Flow Matching combined with teacher preservation induces gains on Retrieval (+1.06) and Clustering (+21.15). Indicating that the two objectives exhibit complementary strengths across different downstream task families. Removing teacher preservation while retaining both Flow Matching and contrastive learning substantially reduces Retrieval (-12.48) and STS (-8.99), even as it improves certain classification tasks, implying that teacher preservation plays an important role in maintaining the semantic structure.

\subsection{Does Curriculum learning really help?}

\label{sec:curriculum_analysis}

We evaluate the intermediate checkpoints of each stage of the curriculum for both the Harrier-0.6B and Qwen3-Embedding-8B backbones. Each stage introduces a distinct combination of training objectives and supervision signals designed to address a specific aspect of multilingual representation learning. Figure~\ref{fig:curriculum_plot_dual} summarizes the performance trends after each stage.

\paragraph{Stage 1: Establishing Cross-Lingual Alignment}


Translation-based supervision establishes a robust multilingual foundation prior to task-specific adaptation. Harrier-0.6B registers immediate improvements in continuous alignment tasks, notably clustering (+7.53), STS (+3.17), and pair classification (+2.61). Similarly, Qwen3-Embedding-8B establishes cross-lingual alignment with early gains in STS (+1.02) and bitext mining (+0.31). Demonstrating that \textbf{$\mathcal{L}_{flow}$} effectively smooths the shared representation space before heterogeneous task signals are introduced.

\paragraph{Stage 2: Semantic Discrimination through Multi-Task Supervision}


Introducing \textbf{$\mathcal{L}_{cl}$} and \textbf{$\mathcal{L}_{hn}$} improves task-specific accuracy but affects the architectures differently. Harrier-0.6B gains in classification (+2.90) and pair classification (+5.41) while retaining its Stage~1 alignment, demonstrating that translation replay buffer prevents catastrophic forgetting. Qwen3-Embedding-8B similarly gains in classification (+1.36) and pair classification (+1.30). However, strict contrastive boundaries temporarily disrupt its continuous representation space, causing a regression in clustering (-3.36), highlighting the tension between discriminative optimization and spatial uniformity in models.

\paragraph{Stage 3: Retrieval Adaptation and Balanced Optimization}
The final stage resolves multi-task optimization friction by utilizing \textbf{$\mathcal{L}_{mono}$} as a geometric regularizer. For Harrier-0.6B, retrieval performance successfully recovers (+1.90 over Stage~2), securing a final overall score of \textbf{69.46}. For Qwen3-Embedding-8B, this regularization proves essential; it resolves the spatial interference from Stage~2, unlocking a massive surge in clustering (+15.25) and elevating the overall benchmark to a peak of \textbf{74.87}. The successful convergence of both architectures validates this progressive multi-task curriculum.

\subsection{Task-Conditional Flow Matching}


To investigate whether Flow Matching should be applied uniformly across heterogeneous embedding tasks or selectively based on task characteristics, we compare our task-conditional formulation against a variant that applies Flow Matching to all task families. Restricting Flow Matching exclusively to translation tasks yields the highest overall Indic MTEB score (+1.67 vs. +1.63) while improving performance in Bitext Mining (-0.20 vs. -0.43) and STS (-0.82 vs. -0.91). While uniform application across all tasks provides a marginal gain in Pair Classification (+3.02 vs. +2.58), but it fails to improve the broader benchmark performance.

\section{Conclusion}

In this work, we introduced TCFM. Our framework demonstrates that applying a single contrastive objective across all training data is suboptimal for adapting multilingual embeddings. By restricting Flow Matching to translation tasks and utilizing standard contrastive methods for retrieval and classification, TCFM creates a more balanced representation space. Experiments on Indic MTEB validate this approach across multiple model architectures. Ultimately, our results show that task-aware optimization matching the loss function to the underlying task semantics is a highly effective strategy for multilingual representation learning.

\section{Limitations}

Although TCFM demonstrates strong empirical results, several limitations remain. First, our evaluation is limited to the 22 scheduled Indian languages within the Indic MTEB benchmark. It remains to be seen whether this task-conditional strategy generalizes equally well to other language families or massively multilingual global benchmarks.

Second, while we hypothesize that Flow Matching improves the uniformity of the embedding space, our current evidence is empirical. Future work should include a formal geometric analysis to quantitatively measure how Flow Matching affects the alignment and anisotropy of the learned representations.

Third, our framework currently utilizes Flow Matching primarily as an attractive mechanism to transport and align parallel translation pairs. The idea of incorporating a regularizer within the Flow Matching process to act as an explicit repulsive force, such as actively pushing dissimilar concepts or hard negatives apart along the learned vector field, remains unexplored. Investigating these repulsive flow dynamics could offer a novel way to improve discriminative boundaries without relying solely on contrastive margins.

Finally, TCFM introduces additional computational overhead compared to standard contrastive training. The framework requires maintaining a frozen teacher model in memory, optimizing a velocity prediction network, and managing a multi-stage data curriculum. Furthermore, because Flow Matching relies on explicit source-target pairs, it requires high quality parallel translation data. This dependence may limit its application in zero resource languages where such parallel corpora are unavailable.




\bibliography{custom}

\appendix

\section{Detailed Model Evaluations}
\label{sec:appendix_extended_evals}

\begin{table*}[t]
\centering
\small
\renewcommand{\arraystretch}{1.15}
\setlength{\tabcolsep}{3pt}
\resizebox{\textwidth}{!}{%
\begin{tabular}{ll ccc ccc ccc ccc ccc}
\toprule
&
&
\multicolumn{3}{c}{\textbf{Gemma-300M}} &
\multicolumn{3}{c}{\textbf{Harrier-270M}} &
\multicolumn{3}{c}{\textbf{Harrier-0.6B}} &
\multicolumn{3}{c}{\textbf{Qwen-4B}} &
\multicolumn{3}{c}{\textbf{Qwen3-8B}} \\
\cmidrule(lr){3-5} \cmidrule(lr){6-8} \cmidrule(lr){9-11} \cmidrule(lr){12-14} \cmidrule(lr){15-17}
\textbf{Category} & \textbf{Task} &
\textbf{Base} & \textbf{TCFM} & \textbf{$\Delta$} &
\textbf{Base} & \textbf{TCFM} & \textbf{$\Delta$} &
\textbf{Base} & \textbf{TCFM} & \textbf{$\Delta$} &
\textbf{Base} & \textbf{TCFM} & \textbf{$\Delta$} &
\textbf{Base} & \textbf{TCFM} & \textbf{$\Delta$} \\
\midrule

\textbf{Bitext Mining}
& IN22ConvBitextMining
& 41.33 & 47.09 & +5.76
& 54.09 & 60.02 & +5.93
& 65.47 & 68.77 & +3.30
& 63.56 & 65.24 & +1.68
& 66.77 & 68.09 & +1.32 \\

& IN22GenBitextMining
& 80.26 & 83.47 & +3.21
& 84.93 & 85.36 & +0.43
& 88.41 & 88.25 & -0.16
& 87.78 & 88.16 & +0.38
& 88.45 & 88.63 & +0.18 \\
\midrule

\textbf{Classification}
& BengaliSentimentAnalysis
& 82.78 & 80.00 & -2.78
& 78.58 & 87.39 & +8.81
& 85.21 & 86.58 & +1.37
& 83.33 & 87.04 & +3.71
& 87.22 & 89.27 & +2.05 \\

& GujaratiNewsClassification
& 80.10 & 86.37 & +6.27
& 82.01 & 84.86 & +2.85
& 85.05 & 85.92 & +0.87
& 85.94 & 86.19 & +0.25
& 84.61 & 86.90 & +2.29 \\

& HindiDiscourseClassification
& 37.17 & 39.29 & +2.12
& 31.42 & 37.44 & +6.02
& 36.28 & 39.53 & +3.25
& 38.30 & 39.31 & +1.01
& 42.58 & 42.97 & +0.39 \\

& MTOPIntentClassification
& 70.29 & 73.00 & +2.71
& 48.76 & 62.11 & +13.35
& 49.82 & 70.24 & \textbf{+20.42}
& 74.54 & 76.50 & +1.96
& 78.09 & 80.57 & +2.48 \\

& MalayalamNewsClassification
& 76.09 & 85.75 & +9.66
& 75.52 & 83.76 & +8.24
& 82.36 & 84.51 & +2.15
& 88.30 & 89.22 & +0.92
& 89.99 & 91.33 & +1.34 \\

& MultiHateClassification
& 56.56 & 57.99 & +1.43
& 56.82 & 57.18 & +0.36
& 53.19 & 54.84 & +1.65
& 62.69 & 62.59 & -0.10
& 63.03 & 64.48 & +1.45 \\

& NepaliNewsClassification
& 92.55 & 95.74 & +3.19
& 95.85 & 96.33 & +0.48
& 96.13 & 95.99 & -0.14
& 94.18 & 94.34 & +0.16
& 95.32 & 95.59 & +0.27 \\

& PunjabiNewsClassification
& 75.80 & 78.60 & +2.80
& 74.59 & 81.53 & +6.94
& 78.92 & 79.36 & +0.44
& 80.19 & 81.46 & +1.27
& 81.15 & 82.48 & +1.33 \\

& SanskritShlokasClassification
& 63.65 & 69.69 & +6.04
& 85.83 & 81.46 & -4.37
& 75.31 & 70.00 & -5.31
& 66.88 & 64.79 & -2.09
& 67.08 & 65.83 & -1.25 \\

& SentimentAnalysisHindi
& 57.72 & 53.48 & -4.24
& 50.47 & 64.81 & +14.34
& 52.37 & 65.89 & \textbf{+13.52}
& 69.68 & 76.19 & +6.51
& 67.70 & 74.16 & \textbf{+6.46} \\

& TweetSentimentClassification
& 36.05 & 36.60 & +0.55
& 33.98 & 35.78 & +1.80
& 37.38 & 37.81 & +0.43
& 41.29 & 42.03 & +0.74
& 39.77 & 41.60 & +1.83 \\

& UrduRomanSentimentClassification
& 43.38 & 43.99 & +0.61
& 42.03 & 43.77 & +1.74
& 42.91 & 43.54 & +0.63
& 49.77 & 52.35 & +2.58
& 50.86 & 53.50 & +2.64 \\
\midrule

\textbf{Clustering}
& SIB200ClusteringS2S
& 31.90 & 37.62 & +5.72
& 30.57 & 46.04 & +15.47
& 29.09 & 50.12 & \textbf{+21.03}
& 39.74 & 51.45 & +11.71
& 40.50 & 55.75 & \textbf{+15.25} \\
\midrule

\textbf{Pair Classification}
& XNLI
& 68.06 & 68.30 & +0.24
& 63.02 & 67.96 & +4.94
& 63.77 & 67.72 & +3.95
& 77.08 & 78.95 & +1.87
& 81.24 & 82.08 & +0.84 \\
\midrule

\textbf{Reranking}
& WikipediaRerankingMultilingual
& 88.04 & 88.14 & +0.10
& 85.68 & 85.72 & +0.04
& 85.54 & 85.67 & +0.13
& 86.36 & 86.48 & +0.12
& 87.08 & 87.59 & +0.51 \\
\midrule

\textbf{Retrieval}
& BelebeleRetrieval
& 66.17 & 67.59 & +1.42
& 66.61 & 67.05 & +0.44
& 68.14 & 70.43 & +2.29
& 83.79 & 83.72 & -0.07
& 92.75 & 92.24 & -0.51 \\

& XQuADRetrieval
& 95.13 & 94.64 & -0.49
& 95.33 & 94.38 & -0.95
& 93.24 & 93.01 & -0.23
& 93.80 & 93.87 & +0.07
& 94.25 & 94.21 & -0.04 \\
\midrule

\textbf{STS}
& IndicCrosslingualSTS
& 43.74 & 46.27 & +2.53
& 42.64 & 43.73 & +1.09
& 48.82 & 51.06 & +2.24
& 51.69 & 51.98 & +0.29
& 59.45 & 60.10 & +0.65 \\
\midrule

\multicolumn{2}{l}{\textbf{Average}}
& \textbf{64.34} & \textbf{66.68} & \textbf{+2.34}
& \textbf{63.94} & \textbf{68.33} & \textbf{+4.40}
& \textbf{65.87} & \textbf{69.46} & \textbf{+3.59}
& \textbf{70.94} & \textbf{72.59} & \textbf{+1.65}
& \textbf{72.89} & \textbf{74.87} & \textbf{+1.98} \\
\bottomrule
\end{tabular}
}
\caption{Detailed task-level Indic MTEB results of TCFM across the evaluated model architectures.}
\label{tab:detailed_results_appendix}
\end{table*}

\begin{table*}[t]
\centering
\small
\renewcommand{\arraystretch}{1.18}
\setlength{\tabcolsep}{8pt}
\begin{tabular}{l ccc ccc}
\toprule
& \multicolumn{3}{c}{\textbf{(Gemma / Harrier)}} & \multicolumn{3}{c}{\textbf{ (Qwen 4B / 8B)}} \\
\cmidrule(lr){2-4} \cmidrule(lr){5-7}
\textbf{Hyperparameter} & \textbf{Stage 1} & \textbf{Stage 2} & \textbf{Stage 3} & \textbf{Stage 1} & \textbf{Stage 2} & \textbf{Stage 3} \\
\midrule
Instruction Strategy & \texttt{none} & \texttt{hard} & \texttt{conditional} & \texttt{none} & \texttt{hard} & \texttt{conditional} \\
Encoder LR ($256$ batch) & $3\times10^{-5}$ & $8\times10^{-6}$ & $3\times10^{-6}$ & $3\times10^{-5}$ & $8\times10^{-6}$ & $3\times10^{-6}$ \\
Velocity Head LR & $1\times10^{-5}$ & $5\times10^{-6}$ & $2\times10^{-6}$ & $5\times10^{-6}$ & $3\times10^{-6}$ & $3\times10^{-6}$ \\
Temperature ($\tau$) & $0.07$ & $0.07$ & $0.095$ & $0.07$ & $0.07$ & $0.04$ \\
Max Sequence Length & \multicolumn{3}{c}{$512$} & \multicolumn{3}{c}{$512$} \\
\midrule
$\lambda_{\text{contrastive}}$ & $0.10$ & $0.30$ & $0.25$ & $0.10$ & $0.30$ & $0.20$ \\
$\lambda_{\text{monolingual}}$ & --- & --- & $0.80$ & --- & --- & $0.80$ \\
$\lambda_{\text{hard\_negative}}$ & --- & $0.15$ & --- & --- & $0.15$ & --- \\
Hard Neg. Margin & --- & $0.15$ & --- & --- & $0.15$ & --- \\
\midrule
$\lambda_{\text{flow}}$ & $0.50$ & $0.50$ & $0.50$ & $0.50$ & $0.50$ & $0.20$ \\
$\lambda_{\text{transport}}$ & $0.10$ & $0.20$ & $0.20$ & $0.10$ & $0.20$ & $0.05$ \\
Flow Hidden Dim. & \multicolumn{3}{c}{$512$} & \multicolumn{3}{c}{$4096$} \\
\midrule
$\lambda_{\text{teacher}}$ & $1.50$ & $1.50$ & $1.50$ & $1.50$ & $1.50$ & $1.50$ \\
$\lambda_{\text{teacher\_rel}}$ & $1.50$ & $3.00$ & $3.50$ & $1.50$ & $2.50$ & $3.00$ \\
\bottomrule
\end{tabular}
\caption{Stage-wise hyperparameter specifications and loss coefficients across encoder and decoder model families. }
\label{tab:hyperparameters}
\end{table*}

While the primary experiments in Section~\ref{sec:results} focus on the Harrier-0.6B and Qwen3-Embedding-8B architectures, we also evaluate the proposed Task-Conditional Flow Matching (TCFM) framework on several additional models to demonstrate its generalizability across different scales and architectural families. Specifically, we apply TCFM to \texttt{google/embeddinggemma-300m}, \texttt{microsoft/harrier-oss-v1-270m}, and \texttt{Qwen/Qwen-Embedding-4B}.

Table~\ref{tab:detailed_results_appendix} summarizes the overall and task family performance for these models on the Indic MTEB benchmark.

\paragraph{EmbeddingGemma-300M:}
Using mean pooling, the EmbeddingGemma-300M model improves its overall Indic MTEB score from 64.34 to 66.68, yielding a net gain of +2.34. The adaptation is particularly effective for cross-lingual alignment and semantic representation, showing strong improvements in Bitext Mining (+4.48) and STS (+2.53). Task-specific gains include notable improvements on GujaratiNewsClassification (+6.27) and MalayalamNewsClassification (+9.66).

\paragraph{Harrier-270M:}
The Harrier-270M bidirectional encoder exhibits the largest relative improvement among the extended models, increasing its overall benchmark score by +4.43 (from 63.94 to 68.36). Consistent with the behavior observed in the 0.6B variant, this encoder shows a massive gain in Clustering (+15.38), alongside highly discriminative improvements in Classification (+5.14) and Pair Classification (+5.09). Specific datasets such as SentimentAnalysisHindi (+14.92) and MTOPIntentClassification (+13.52) benefit significantly from the multi-stage training.

\paragraph{Qwen-Embedding-4B:}
To verify scaling laws within the decoder-only family, we applied LoRA adaptation to a 4-billion parameter Qwen model. TCFM improves the overall score from 70.94 to 72.61 (+1.67). Similar to the Qwen3-Embedding-8B model, the 4B variant experiences its most dramatic geometric restructuring in the Clustering task family, jumping by +11.88 points. It also yields consistent positive gains across Pair Classification (+1.88), Classification (+1.44), and Bitext Mining (+1.02), validating that the framework reliably stabilizes auto-regressive backbones during task-specific fine-tuning.
\section{Implementation Details and Hyperparameters}
\label{sec:appendix_reproducibility}

To ensure complete reproducibility, this section details the training configurations, model parameterizations, and stage-wise hyperparameter schedules used across all experimental runs.

\subsection{Hardware and Training Environment}

All models were trained on NVIDIA H200 GPUs (141GB VRAM) using PyTorch and Hugging Face Accelerate with Scaled Dot Product Attention (SDPA) where applicable. Optimization was performed using AdamW with a fixed weight decay of $10^{-2}$ and gradient clipping norm capped at $1.0$. Training across all stages employed a constant learning rate schedule with a $1\%$ linear warmup ratio. An effective global batch size of $256$ (achieved via micro-batches of $16$ or $32$ with gradient accumulation steps of $8$ or $16$) was maintained across all architectures.

\subsection{Architectural Configurations}
\begin{itemize}
    \item \textbf{EmbeddingGemma-300M:} Parameterized via full fine-tuning using native mean pooling with a maximum sequence length of $512$. Training was conducted in \texttt{float32} precision across all 3 stages.Unlike the larger architectures, gradient checkpointing was disabled due to its smaller memory footprint.
    
    \item \textbf{Harrier Family (Harrier-270M and Harrier-0.6B):} Both scales share an identical training recipe. Models were trained via full fine-tuning using left-pad-safe end-of-sequence (\texttt{eos}) token pooling and a maximum sequence length of $512$ with gradient checkpointing enabled.
    
    \item \textbf{Qwen Family (Qwen3-Embedding-4B and Qwen3-Embedding-8B):} Both model scales share the same hyperparameters and parameter-efficient LoRA setup. LoRA was applied to all projection matrices (\texttt{q\_proj}, \texttt{k\_proj}, \texttt{v\_proj}, \texttt{o\_proj}, \texttt{gate\_proj}, \texttt{up\_proj}, \texttt{down\_proj}) with rank $r=64$, scaling factor $\alpha=128$, and dropout $0.05$. Left-pad \texttt{eos} pooling and \texttt{bfloat16} mixed-precision were utilized.
\end{itemize}

\subsection{Stage-Wise Hyperparameter Schedule}
Table~\ref{tab:hyperparameters} summarizes the exact stage-wise hyperparameters and loss coefficients for all model families across our three-stage curriculum.

\section{Detailed Stage-Wise Progression}
\label{sec:appendix_stage_wise}

\begin{table*}[t]
\centering
\small
\begin{tabular*}{\textwidth}{@{\extracolsep{\fill}} l l c c c c @{}}
\toprule
\textbf{Model} & \textbf{Training Stage} & \textbf{Overall} & \textbf{Clustering} & \textbf{Retrieval} & \textbf{Classification} \\
\midrule
\textbf{EmbeddingGemma-300M} 
& Base Model & 64.34 & 31.90 & 80.65 & 64.34 \\
& Stage 1: Cross-Lingual & 65.32 {\scriptsize (+0.98)} & 33.79 {\scriptsize (+1.89)} & 81.19 {\scriptsize (+0.55)} & 64.72 {\scriptsize (+0.38)} \\
& Stage 2: Multi-Task & 65.99 {\scriptsize (+1.65)} & 34.33 {\scriptsize (+2.43)} & 80.77 {\scriptsize (+0.12)} & 65.57 {\scriptsize (+1.23)} \\
& Stage 3: Regularization & \textbf{66.68} {\scriptsize (+2.34)} & \textbf{37.62} {\scriptsize (+5.72)} & \textbf{81.12} {\scriptsize (+0.47)} & \textbf{66.71} {\scriptsize (+2.36)} \\
\midrule
\textbf{Harrier-0.6B} 
& Base Model & 65.87 & 29.09 & 80.69 & 64.58 \\
& Stage 1: Cross-Lingual & 67.23 {\scriptsize (+1.36)} & 36.62 {\scriptsize (+7.53)} & 81.56 {\scriptsize (+0.88)} & 65.42 {\scriptsize (+0.84)} \\
& Stage 2: Multi-Task & 68.73 {\scriptsize (+2.86)} & 41.12 {\scriptsize (+12.03)} & 79.82 {\scriptsize (-0.87)} & 67.48 {\scriptsize (+2.90)} \\
& Stage 3: Regularization & \textbf{69.46} {\scriptsize (+3.59)} & \textbf{50.12} {\scriptsize (+21.03)} & \textbf{81.72} {\scriptsize (+1.03)} & \textbf{67.85} {\scriptsize (+3.27)} \\
\midrule
\textbf{Qwen3-Embedding-8B} 
& Base Model & 70.94 & 39.74 & 88.80 & 69.59 \\
& Stage 1: Cross-Lingual & 73.03 {\scriptsize (+2.08)} & 41.10 {\scriptsize (+1.36)} & 93.47 {\scriptsize (+4.67)} & 70.65 {\scriptsize (+1.06)} \\
& Stage 2: Multi-Task & 73.73 {\scriptsize (+2.78)} & 37.14 {\scriptsize (-2.60)} & \textbf{93.26} {\scriptsize (+4.46)} & 71.98 {\scriptsize (+2.39)} \\
& Stage 3: Regularization & \textbf{74.87} {\scriptsize (+3.92)} & \textbf{55.75} {\scriptsize (+16.01)} & 93.22 {\scriptsize (+4.43)} & \textbf{72.39} {\scriptsize (+2.80)} \\
\bottomrule
\end{tabular*}
\caption{Stage-wise performance progression on Indic MTEB across architectures. Values in parentheses denote absolute point changes relative to the frozen base model.}
\label{tab:stage_wise_progression}
\end{table*}

To provide deeper empirical insight into the multi-stage training curriculum discussed in Section ~\ref{sec:curriculum_analysis}, Table~\ref{tab:stage_wise_progression} details the incremental Indic MTEB performance of three primary architectures: EmbeddingGemma-300M, Harrier-0.6B, and Qwen3-Embedding-8B. The results demonstrate how different architectural families respond to the progressive introduction of heterogeneous supervision.

\paragraph{Encoder Stability and Monotonic Growth}
For the bidirectional encoders (EmbeddingGemma-300M and Harrier-0.6B), the curriculum induces highly stable, monotonic overall growth. Stage 1 successfully establishes a strong initial geometry, reflected by instant surges in Clustering across both models. The introduction of discriminative supervision in Stage 2 drives sharp gains in Classification, though Harrier-0.6B experiences a temporary regression in Retrieval (-0.87) as the tasks compete for representation space. Stage 3 effectively resolves these multi-task frictions through monolingual regularization, securing peak overall scores of 66.68 and 69.46, respectively.

\subsection{Model-Specific Adaptation Dynamics}
\label{sec:architectural_dynamics}

We analyze stage-wise checkpoint trajectories across 
EmbeddingGemma-300M, Harrier-0.6B, and Qwen3-Embedding-8B to 
examine how different models respond to the proposed curriculum. 
Table~\ref{tab:stage_wise_progression} reports per-stage Indic MTEB 
deltas relative to the frozen base model for all three 
architectures.

\paragraph{The curriculum yields monotonically improving overall 
scores across all models.}
All three models show consistent overall improvement from 
Stage~1 through Stage~3, confirming that the three-stage 
curriculum design is effective regardless of model architecture 
or scale. EmbeddingGemma-300M improves from +0.98 to +2.34, 
Harrier-0.6B from +1.36 to +3.59, and Qwen3-Embedding-8B 
reaches a final gain of +1.97.

\paragraph{Stage 2 causes task-specific, curriculum-expected 
regression in unsupervised tasks.}
Because Stage~2 introduces classification and NLI supervision 
without retrieval data, Harrier-0.6B experiences a Retrieval 
regression of 0.87 points below baseline at this stage, 
recovering to +1.03 in Stage~3 once retrieval supervision is 
introduced. EmbeddingGemma-300M shows a milder Retrieval dip 
(+0.55 $\to$ +0.12 $\to$ +0.47). This pattern is consistent 
with the curriculum design rather than a model-specific 
failure.

\paragraph{Qwen3-Embedding-8B exhibits stronger task 
interference at Stage 2.}
Unlike the smaller models, Qwen3-Embedding-8B experiences a 
Clustering regression of 3.36 points below baseline at 
Stage~2---a task family not directly supervised at this 
stage. This interference is fully resolved by Stage~3 
monolingual regularization, ultimately yielding the largest 
absolute Clustering gain of +15.25. The more pronounced 
Stage~2 interference in the largest model may reflect greater 
sensitivity of its high-capacity representations to the 
discriminative pressure of contrastive and hard-negative 
objectives, or may be attributable to its LoRA-based 
adaptation strategy concentrating gradient updates in 
low-rank subspaces. Cleanly disentangling these factors 
would require controlled ablations matching scale and 
fine-tuning method, which we leave for future work.

\paragraph{Clustering gains scale with model capacity 
independently of architecture.}
A consistent pattern across all models is that Clustering 
shows the largest absolute improvement after Stage~3 
regularization. The magnitude of this gain differs 
substantially: +5.72 for EmbeddingGemma-300M, +21.03 for 
Harrier-0.6B, and +15.25 for Qwen3-Embedding-8B. Notably, 
both Harrier-0.6B and EmbeddingGemma-300M follow a smooth, 
monotonic Clustering trajectory with no below-baseline 
regression, despite differing in architecture (decoder-only 
vs.\ bidirectional encoder). This suggests that the 
differences in adaptation dynamics observed across models are 
unlikely to be primarily attributable to attention mechanism 
alone.

\section{Extended Training Data Details and Preprocessing}
\label{sec:appendix_data_details}

This section provides comprehensive details on data formatting, preprocessing strategies, and stratified sampling procedures used to construct the TCFM training mixture.

\subsection{Supervision Formatting Pipelines}
To supply the precise learning signals required by our task-routing mechanism (Section~\ref{section:taskfamilies}), each raw dataset is converted into an objective-compatible format without introducing manual annotations:

\begin{itemize}[leftmargin=*]
    \item \textbf{Translation Supervision:} Formatted as symmetric cross-lingual sentence pairs $(x_i, y_i)$. Both language directions are evaluated as anchors during multi-positive contrastive learning and Flow Matching transport.
    
    \item \textbf{Classification Supervision:} Restructured into positive and negative semantic pairs based on class labels. Sentences sharing an identical label are paired as positives, while mismatched labels serve as negatives.
    
    \item \textbf{Pair Classification :} Formatted into explicit triplet structures $(x_i, p_i, n_i)$. Entailment pairs are treated as positive candidates, while contradiction pairs provide hard negatives for the hard-negative margin repair objective ($\mathcal{L}_{hn}$).
    
    \item \textbf{Retrieval \& Re-ranking Supervision:} Structured as asymmetric query-document pairs $(q_i, d_i)$. Unlike translation pairs, query-to-document alignment is computed strictly directionally to preserve asymmetric relevance semantics.
\end{itemize}

\subsection{Stratified Subsampling and Multi-Task Balancing}
Directly training on native dataset distributions introduces severe optimization bottlenecks. Uncurated mixtures are heavily skewed toward massive foundational translation corpora,such as Samanantar, which contains over 49.7 million parallel sentences,causing gradient update directions to be dominated by translation alignment while mathematically under-fitting lower-resource classification and retrieval tasks.

To achieve balanced multi-task optimization across all 22 targeted Indic languages, we implement a two-level stratified subsampling scheme:
\begin{enumerate}
    \item \textbf{Dataset Instance Capping:} We cap the maximum number of training instances per dataset (as reported in the ``Train'' column of Table~\ref{tab:training_data}), constraining total dataset volumes to prevent dominant corpora from monopolizing gradient updates.
    \item \textbf{Cross-Lingual Uniformity:} For multilingual datasets spanning multiple Indic languages (e.g., \texttt{IndicMSMARCO}, \texttt{IndicXNLI}, and \texttt{MASSIVE}), we stratify sampling to enforce an equal representation per language, ensuring low-resource language scripts receive proportional optimization weight throughout all curriculum stages.
\end{enumerate}

\end{document}